\documentclass[12pt]{spieman}  
\usepackage{amsmath,amsfonts,amssymb}
\usepackage{graphicx}
\usepackage{setspace}
\usepackage{tocloft}
\usepackage[toc,page]{appendix}
\usepackage{lineno}
\usepackage{xcolor}
\title{White Box Methods for Explanations of Convolutional Neural Networks in Image Classification Tasks}

\author[a,*]{Meghna P Ayyar}
\author[a]{Jenny Benois-Pineau}
\author[a]{Akka Zemmari}

\affil[a]{University of Bordeaux, LaBRI, UMR 5800, 351, Crs de la Liberation, 33405,
Talence, France}

\cftpagenumbersoff{figure}
\cftpagenumbersoff{table} 
\begin{document} 
\maketitle

\begin{abstract}
In recent years, deep learning has become prevalent to solve applications from multiple domains. Convolutional Neural Networks (CNNs) particularly have demonstrated state of the art performance for the task of image classification. However, the decisions made by these networks are not transparent and cannot be directly interpreted by a human. Several approaches have been proposed to explain to understand the reasoning behind a prediction made by a network. In this paper, we propose a topology of grouping these methods based on their assumptions and implementations. We focus primarily on white box methods that leverage the information of the internal architecture of a network to explain its decision. Given the task of image classification and a trained CNN, this work aims to provide a comprehensive and detailed overview of a set of methods that can be used to create explanation maps for a particular image, that assign an importance score to each pixel of the image based on its contribution to the decision of the network. We also propose a further classification of the white box methods based on their implementations to enable better comparisons and help researchers find methods best suited for different scenarios.



\end{abstract}


{\noindent \footnotesize\textbf{*}Meghna P Ayyar,  \linkable{meghna-parameswaran.ayyar@etu.u-bordeaux.fr}}

\begin{spacing}{2}   


\section{Introduction}
\label{sect:intro}  
Deep learning approaches which are a part of methods we call today Artificial Intelligence (AI), have become indispensable for a wide range of applications requiring analysis and understanding of a large amount of data. They can produce promising results that outperform human capabilities in various decision tasks relating to visual content classification and understanding such as face detection\cite{garcia2004convolutional}, object detection and segmentation\cite{RomanBDPCR17}, image denoising\cite{wen2021denoising}, video-based tasks like sports action recognition \cite{MartinBPM20} and saliency detection\cite{shang2021moving} amongst others. The success of deep learning-based systems in these tasks have also paved the way for their applications to be developed for a variety of medical diagnosis tasks like cancer detection \cite{wu2019deep}, and Alzhiemer's disease detection\cite{aderghal2020improving} on different imaging modalities just to name a few. Along with the usefulness of these tools, the trustfulness and reliability of such systems is also being questioned. 

Though the results of deep learning models have been exemplary they are not perfect, can produce errors, are sensitive to noise in data and often lack the transparency to have verifiability of the decisions that they make. A specific example is related to the visual task of object classification from an image. The study by Ribeiro et al.\cite{ribeiro2016should} showed that the trained network that performed supervised image classification, used the presence of snow as the distinguishing feature between the \textit{"Wolf"} and \textit{"Husky"} named classes present in the dataset. Such limitations raise ethical and reliability concerns that need to be addressed before such systems can be deployed and adopted on a wider scale. The objective of explainable AI/Deep learning is to design and develop methods that can be used to understand how these systems produce their decisions.
\begin{figure}
    \centering
    \includegraphics[width=0.85\linewidth]{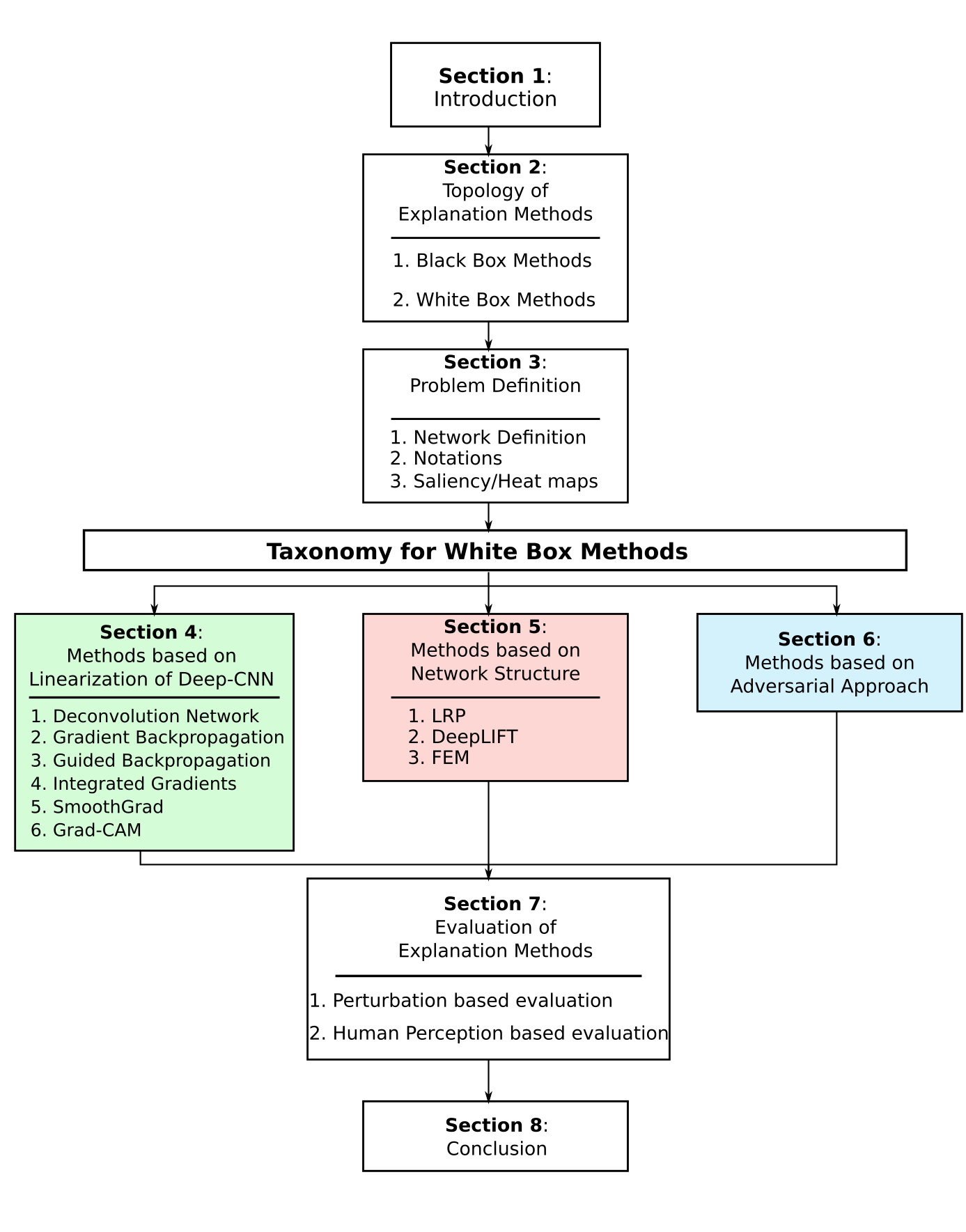}
    \caption{Organization of the Paper}
    \label{fig:paper_org}
\end{figure}

The behaviour described in the case of the wolf/husky classification has been termed as the problem of a trained classifier behaving like a "Clever Hans" predictor\cite{lapuschkin2019unmasking}. Explanation methods aid in the unmasking of such spurious correlations and biases in the model or data and also in understanding of the failure cases of the system. If we can comprehend the reasoning behind the decision of a model, it could also help in uncovering associations that had been previously unobserved, which could aid in furthering the future research trends. It is important to mention  that explainability  focuses on the attribution of the output based on the input. It does not deal with answering the causality of the features or factors that have lead to a decision that has been taken. That is the explainers are only correlation-based (input-output) and do not make causal inferences.

The current study focuses on the task of supervised image classification using specific deep neural network architectures ,i.e. Convolutional Neural Networks (CNNs). CNNs have become one of the most successful architectures concerning AI tasks relating to images. Hence the methods presented in the subsequent sections are focused on finding the relation between the predicted output classes and the input features that are the pixels of the image. We specifically focus on white-box methods, which we found quite promising\cite{ayyar2021icip}. We also give our viewpoint on the topology of the different explanation methods which have been developed up to now. We present the detailed problem definition in Sec.~\ref{sec:prob_def}. The following sections present the different explanation methods in detail. In Sec.~\ref{sec:comparison} we discuss different ways of evaluation of explainers and present the comparison of several of them used for the explanation of a well-known VGG-16 CNN \cite{simonyan2014very} for an image classification task.To facilitate reading of this survey we present the paper organization in Fig. \ref{fig:paper_org}.




\section{Topology of Explanation Methods for Image Classification Tasks} \label{sec:topology}
In the book by Samek et al.\cite{samek2019explainable}, the authors present recent trends in the research in explainable AI and some of the directions for future explorations. They have presented a topology for the various explanations methods like meta-explainers, surrogate/sampling-based, occlusion based and propagation-based to name a few. However, with the addition of newer methods and their adaptations to different types of neural networks and datasets, we propose to update the topology based on the domain to which the methods are applied and their inherent design. Comparing recent studies, two major types of explanation methods exist i) Black-box methods and ii) White Box methods. In this review, for both cases, we mainly focus on the explanations of decisions of trained Deep Neural Network (DNN) classifiers. This means that for each sample of the data the methods we review explain the decision of the network. This is why they are called "sample-based" methods\cite{petkovic2020}.  In the following, we will briefly explain the "Black box" methods and focus on "White box"  methods in image classification tasks.

\subsection{Black Box  Methods} \label{black_box_methods}
\textit{Black Box} refers to an opaque system. The internal functioning of the model is kept hidden or is not accessible to the user. Only the input and the output of the system are accessible and such methods are termed as black box methods as they are model agnostic.

There are multiple ways to examine what a black-box model has learned\cite{guidotti2018survey}. A prominent group of methods are focused on explaining the model as a whole by approximating the black-box model like the neural networks with an inherently interpretable model. One such example is the use of decision trees\cite{krishnan1999extracting}. Decision trees are human-interpretable as the output is based on a sequence of decisions starting from the input data. To approximate a black-box trained network Frosst et al.\cite{FrosstH17} have used multiple input-output pairs generated by the network to train a soft decision tree that could mimic the network's behaviour. Each node makes a binary decision and learns a filter $w$ and a bias $b$ term and the probability of the right branch of the tree being selected is given by Eq.~(\ref{decision_tree}) where the function $\sigma(x) =  \frac{1}{1 + e^{-x}} $ is the sigmoid logistic function, $x$ is the input and $i$ is the current node.

\begin{equation}
    p_i(x) = \sigma(x w_i + b_i)
    \label{decision_tree}
\end{equation}

The leaf nodes learn a simple distribution $Q$ for the different $k$ classes present in the dataset.
This method can be qualified as a Dataset-based explanation, as the decision tree is built for the whole dataset of pairs input-output. 

Sample-based black-box methods deal with explaining a particular output of the model. These methods are not focused on understanding the internal logic of the model for all the classes on a whole but are restricted to explaining the prediction for a single input. The \textit{Local Interpretable Model-agnostic Explanations}\cite{ribeiro2016should} (LIME) method is one such approach that derives explanations for individual predictions. It generates multiple perturbed samples of the same input data and the corresponding outputs from the black box and trains an inherently interpretable model like a decision tree or a linear regressor on this combination to provide explanations.

\begin{figure}
    \centering
    \includegraphics[width=\textwidth]{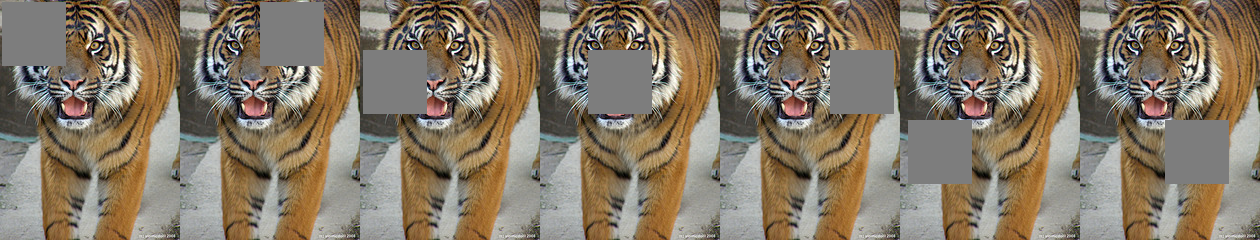}
    \caption{Gray-value sliding window used to occlude to different parts of the image. Image taken from ImageNet \cite{imagenet_cvpr09}}
    \label{fig:occlusion}
\end{figure}

Taking into account human understanding of visual scenes, such as attraction by meaningful objects in visual understanding tasks, for image classification tasks the regions of the image where the objects are present should have a higher contribution to the prediction. Based on this logic, some methods attempt to occlude different parts of the image iteratively using a sliding window mask\cite{zeiler2014visualizing}. Figure~\ref{fig:occlusion} illustrates how the gray-valued window is used for occluding different parts of the images by sliding it across the image. By observing the change in the prediction of the classifier when different regions are hidden, the importance of regions for the final decision is calculated.

Fong et al.\cite{fong2019explanations} also build the explanations on which region contributes to the DNN decision the most by masking. Instead of using a constant grey-value mask, they formulate the explanation as a search for the minimal mask which changes the classification score for the given image the most. The mask applies a meaningful transformation which models image acquisition process like blurring. 
They find the mask by minimizing the expectation of output classification score of the network on the image perturbed using the blurred mask. Instead of using a single mask to perform the search, they apply the perturbation mask stochastically to the image. Also, L1 and total variation regularization are used to ensure that the final mask deletes the smallest subset of the image and has a regular structure.

Nevertheless, these methods only aid in identifying if the network is predicting based on a non-intuitive region in the image. The explanations are not useful to identify which layers or filters in a DNN classifier cause these wrong correlations between the input image regions and the prediction. Thus they cannot be used to improve the network performance. Hence the white box methods, which allow for analyzing internal layers of the network are more interesting.

\subsection{White Box Methods}
The term "white box"  implies a clear box that symbolizes the ability to see into the inner workings of the model i.e. its architecture and the parameters. Due to the extensive research in DNNs, they are not unknown architectures anymore and studies like Yosinski et al.\cite{yosinski2014transferable} have been able to show the types of features that are learnt at the different layers in a DNN. Therefore, multiple methods aim to exploit the available knowledge of the network itself to create a better understanding of the prediction and the internal logic of the network thus allowing for further optimization of the architecture and hyperparameters of the model.

In this overview, we propose to deal with the specific case of Deep Neural network  classifiers such as convolutional neural networks (CNNs). There has been an abundance in the methods proposed to explain the decisions of the CNNs and we need to have a systematic way to compare and understand how they provide the explanations. To do that we propose the following taxonomy for existing "white-box" methods based on their approach used for generating explanations: i) Methods based on Linearization of the Deep-CNN, ii) Methods based on network structure, iii) Methods based on Adversarial approach. Due to the exploding research in the field, we do not claim to be complete in our taxonomy but believe to have addressed the main trends. The main idea is to have a compact grouping of the different methods so that they can be studied based on the similarity of their approach, while also presenting a new group for a method that has a very different approach when compared to others e.g. the adversarial methods. We present in detail the common characteristics of the methods grouped into a category in further sections.

\begin{figure}
    \centering
    \includegraphics[width=\linewidth]{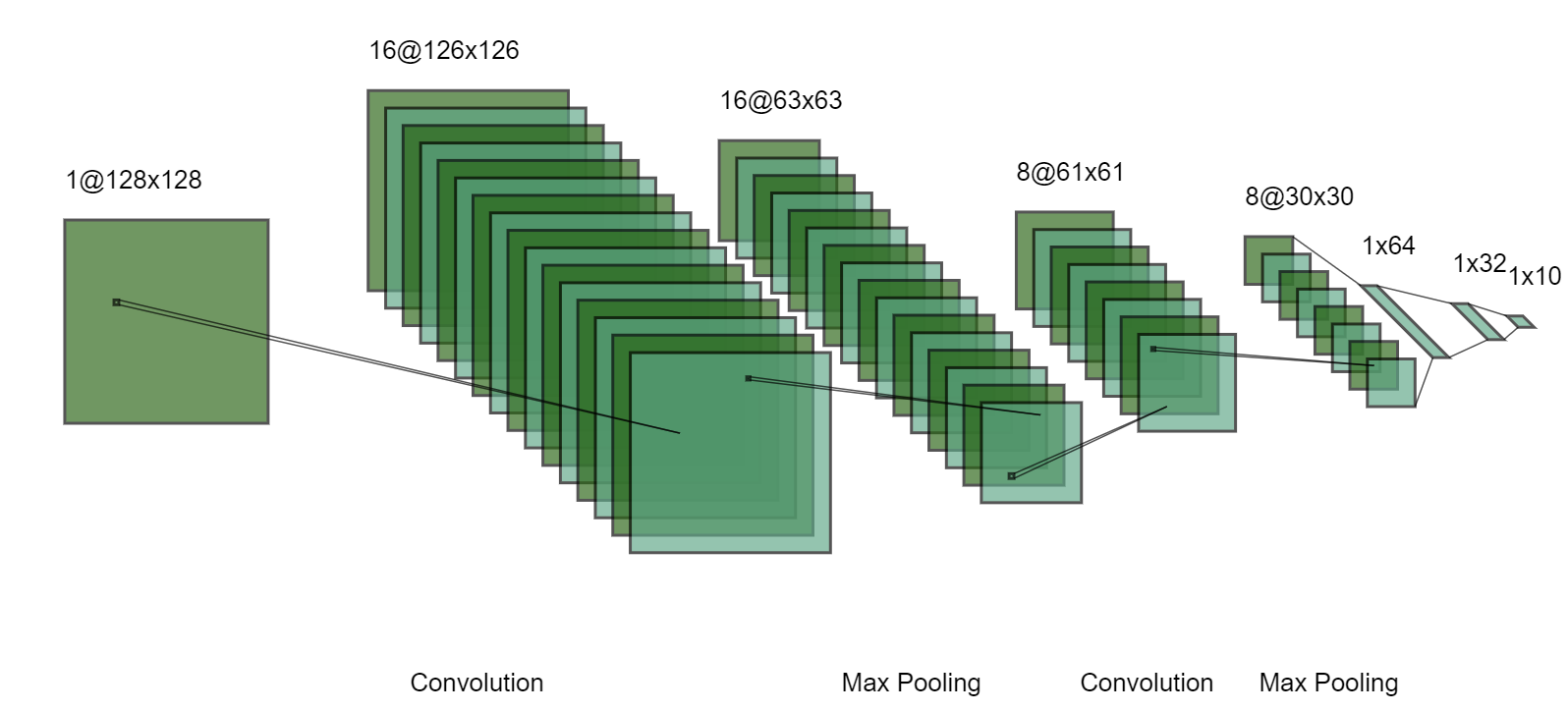}
    \caption{Architecture of a standard CNN: convolutional layers with non-linear activations, pooling layers and a perceptron at the end for classification}
    \label{fig:cnn_diag}
\end{figure}


\section{Problem Definition} \label{sec:prob_def}
This section provides the basic terminology and the definitions required to understand the type of network that we will be focusing on, the notations used and how the results are to be visualized. 

\subsection{Network Definition}
\begin{figure}
    \centering
    \includegraphics[width=0.38\linewidth]{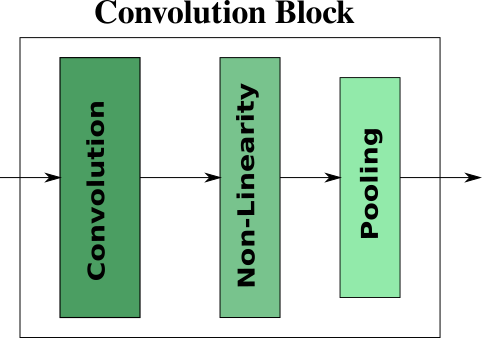}
    \caption{Structure of a convolution block as proposed by Goodfellow et.al.\cite{goodfellow2016deep}}
    \label{fig:cnn_block}
\end{figure}

The problem under consideration is the image classification task. To define the task, first consider a convolutional neural network (CNN). A simple AlexNet-like\cite{krizhevsky2012imagenet} CNN is illustrated in Fig.~\ref{fig:cnn_diag}. The network consists of a series of convolutional layers, a non-Linear activation layer and a pooling layer that form the convolution (conv) block as illustrated by Fig.~\ref{fig:cnn_block}. The conv blocks are followed by fully connected layers (FC) that are simple feed-forward neural networks\cite{goodfellow2016deep}. The Rectified Linear Unit (ReLU) activation shown in Eq.~(\ref{eq:relu}) and Max Pooling are the most commonly used while building  CNN classifiers\cite{zemmari2020deep}. The last layer of the network has the same dimension as the number of classes in the problem, in the example in Fig.~\ref{fig:cnn_diag} it is 10 implying there are 10 categories of objects to recognize. 
\begin{equation}
    ReLU(t) = \max(0,t)
    \label{eq:relu}
\end{equation}
\subsection{Notations}
Consider a CNN that takes as input an image $x$ of size $N \times N$ expressed as $x = [x_{11}, ..., x_{NN}] \in \mathbb{R}^{N \times N} $ and the output of the classification is a C-dimensional vector $S(x) = [S_1(x),..., S_C(x)] \in \mathbb{R}^C$. Here $C$ represents the number of classes and the image $x$ represents the input features for the network. The scores $S_c(x)$ would be the output classification score for the image $x$ for the class $c$. The network thus models a mapping $f: \mathbb{R}^{N \times N} \rightarrow \mathbb{R}^C$. The output score vector is usually normalized to approximate the probability, thus $\mathbb{R}^C$ is restricted to the $[0,1]$ interval and the score's vector $S(x)$ sums to $1$.
The problem of explanations consists in assigning,  to each pixel $x_{ij}, i=1, ..., N, j=1, ..., N$,  a relevance score  $R^c_{ij} \in [0,1]$ with respect to its contribution to the output $S_c(x)$.
Otherwise to produce a relevance score map $R^c = [R^c_{11}, ..., R^c_{NN}] \in \mathbb{R}^N$ for each of the pixels and/or features of internal convolutions layers of the network to the output $S_c(x)$ where the class $c$ can either be the correct label class or a different class where it can be used to analyze the cause for that classification. 

To "explain" pixel importance to the user a visualization of the scores in $R^c$ is usually performed by computing  "Saliency/ Heat maps" and superimposing them on the original image.

\subsection{Saliency/ Heat maps}
\begin{figure}
    \centering
    \includegraphics[width=0.55\linewidth]{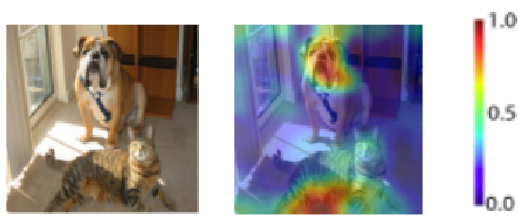}
    \caption{A saliency map visualization for the sample image. Sample image taken from ImageNet dataset\cite{imagenet_cvpr09}}
    \label{fig:heatmap}
\end{figure}

A saliency/heat map is the visualization of the relevance score map $R^c$ with colour Look-up-Tables (LuTs) mapping $[0,1]$ onto a colour scale from blue to red as illustrated by Fig.~\ref{fig:heatmap}. This form of visualizations is necessary for the user to understand and glean insights from the results of the explanation methods. In the current illustration, we have used the "jet" colour map that has a linear transition from the maximum value mapping to red, the middle to yellow-green colour, and the lowest to blue. Other LuTs can also be used for the visualization of the heat maps but we have chosen "jet" as it is one of the more popular colour maps and is intuitively understandable for a human observer.

Given this kind of network classifier and problem formulation, several methods have been proposed that can be employed for the visualization of relevance score maps given a particular image. 


\section{Methods based on Linearization of the Deep-CNN} \label{sec:linear}

A (Convolutional) Neural Network is a non-linear classifier. It can be defined as a mapping  $f(x)$ from the input (feature space) $X$ to the output score space $S$. The methods based on the linearization of a CNN  produce explanations approximating the non-linear mapping $f(x)$. One of the commonly used approximations is the linear approximation 


\begin{equation}
    S_c(x) \approx w^Tx + b 
    \label{eq:fot}
\end{equation}
where the $w$ are the weights, $x$ is the input and $b$ is the bias related to the network. Different methods employ different ways to calculate the weight and bias parameters of the network approximation and thus produce different explanations.

Methods grouped under this category primarily deal with the calculation of gradients of the network or are equivalent in order to perform the linear approximation of the network. They differ in the layers at which the gradients are computed for example gradient backpropagation does it at the input layer whereas Grad-CAM does it at the last convolution layer. The following section presents some of these methods with their intuitions and implementation details.

\subsection{Deconvolution Network based method} \label{sec:deconv_net}
The Deconvolution Network (DeconvNet) proposed by Zeiler et al.\cite{zeiler2014visualizing} was a network that reversed the mapping of a CNN. It builds a mapping of the output score $S_c$ to the space of the input pixels $x_{ij}$. It does not require retraining and directly uses the learned filters of the CNN. Starting with the input image $x$, a full forward pass through the CNN is done to compute the feature activation throughout the layers. To visualize the features of a particular layer the corresponding feature maps from the CNN layer are passed onto the DeconvNet. In the DeconvNet the three steps i) Unpooling ii) Rectification iii) Filtering are done at each layer iteratively till we reach the input features layer. 
\begin{itemize}
    \item Unpooling: Max Pooling operation in a CNN is non-invertible, Hence to reverse this, during the forward pass in the network, at each layer the maxima locations are saved to a matrix called the \textit{Max location switches}. During the unpooling, the values from the previous layers are mapped to only the locations of maxima and the rest of the positions have a 0 assigned to them.
    \item Rectification: After applying the unpooling, a ReLU function (Eq.~\ref{eq:relu}) is applied onto the matrix to ensure that only positive influences on the output are backpropagated.
    \item Filtering: This operation is the inverse operation to the convolution in the forward pass. To achieve this the DeconvNet convolves the rectified maps with the vertically and horizontally flipped version of the filter that has been learned by that layer in the CNN. The authors show that filters thus defined from learnt CNN filters are the deconvolution filters. We show the mathematical derivation of this in Appendix \ref{deconv_grad}.
\end{itemize}
Performing these operations iteratively from the layer of our choice to the input pixel layer helps to reconstruct the features from the layer that correspond to different regions in the input image $x$. The importance of pixels is then expressed with a heat map. 

\subsection{Gradient Backpropagation} \label{sec:grad_backprop}
The gradient backpropagation method\cite{simonyan2013deep} was proposed to explain prediction of a model based on its locally evaluated gradient. The local gradient of the output classification score $S_c$ with respect to the input $x$ at a particular image $x_0$ is used to calculate the weights parameter $w$ from Eq.~(\ref{eq:fot}). This means that the linear approximation of the non-linear mapping $f(x)$ is formulated as a Taylor First Order Expansion of $f(x)$ in the vicinity of a particular image $x_0$, and  $b= f(x_0)$. The weight parameters are thus calculated as in Eq.~(\ref{eq:weights}).

\begin{equation}
    w = \frac{\partial S_c}{\partial x} |_{x_0}
    \label{eq:weights}
\end{equation}

The partial derivative of the output classification score with respect to the input corresponds to the gradient calculation for a single backpropagation pass for a particular input image $x$. It is equivalent to the backpropagation step that is performed during training which usually corresponds to a batch of images. For this method, the notation of gradient at $x_0$ is to show that the backpropagation is for just one image. Also, during the training of a CNN, the backpropagation step stops at the second layer of the network for efficiency as the aim is not to change the input values. But with this method, the backpropagation is performed till the input layer to inspect which pixels affect the output the most.

The final heat map relevance scores $R_{ij}$ for a particular pixel $i,j$ in the input 2D image are calculated as shown in Eq.~(\ref{eq:sensitivity}) in the case of a gray-scale image. 
\begin{equation}
    R_{ij} = \left \| \frac{\partial}{\partial x_{ij}}S_c \right \| 
    \label{eq:sensitivity}
\end{equation}
For an RGB image, the final map is calculated as the maximum weight of that pixel from the weights matrices from each of the three channels as shown in Eq.~(\ref{eq:max}), where $k$ corresponds to the different channels in the image. 

\begin{equation}
    R_{ij} =  \max_k |w(i,j,k)|
    \label{eq:max}
\end{equation}

Also, these gradients can be used for performing a type of sensitivity analysis. The magnitude of the derivatives that have been calculated could be interpreted to indicate the input pixels to which the output classification is the most sensitive. A strong gradient magnitude value would correspond to the pixels that need to be changed the least to affect the final class score the most. 

Simonyan et al.\cite{simonyan2013deep} have also shown that the gradient backpropagation is a generalization of  DeconvNet (Sec.~\ref{sec:deconv_net}). Indeed, this can be shown by comparing the three operations that DeconvNet performs with the gradient calculation.
\begin{itemize}
    \item \textit{Unpooling}: During basic backpropagation at a max-pooling layer the gradients are backpropagated to only those positions that had the max values during the forward pass. This is exactly the same operation that is achieved by the use of the \textit{Max location switches} matrix in the DeconvNet, see Sec.~\ref{sec:deconv_net}.
    \item \textit{Rectification}:
For a CNN with the output of a convolution layer $n$ as $Y$ the application of the ReLU activation is performed as $Y_{n+1} = \max(Y_n , 0)$, where $Y_{n+1}$ then is the input for the next layer in the network. During the gradient backpropagation the rectification applied on the gradient map is based on the input, i.e. on those position where $Y_n > 0$. Whereas, in the DeconvNet the rectification is applied on the unpooled maps and hence corresponds to the condition of $Y_{n+1} > 0$. Figures.~\ref{fig:guid_for_relu}(b) and \ref{fig:guid_for_relu}(c) show the changes in the calculation of the two matrices based on this difference in the operations. 

\item \textit{Filtering}: As shown in Appendix \ref{deconv_grad}, the vertical and horizontally flipped filter that is used during the filtering step in the DeconvNet corresponds to the gradient calculation of the convolution with respect to the input $x$. This is the same step as the gradient backpropagation method performs and hence this step is equivalent for the two methods.
\end{itemize}
Except for the rectification step, the two methods are equivalent in their calculations and therefore, the gradient backpropagation method can be seen as a generalization of the DeconvNet.

\subsection{Guided Backpropagation} 
\begin{figure}
    \centering
    \includegraphics[width=0.85\linewidth]{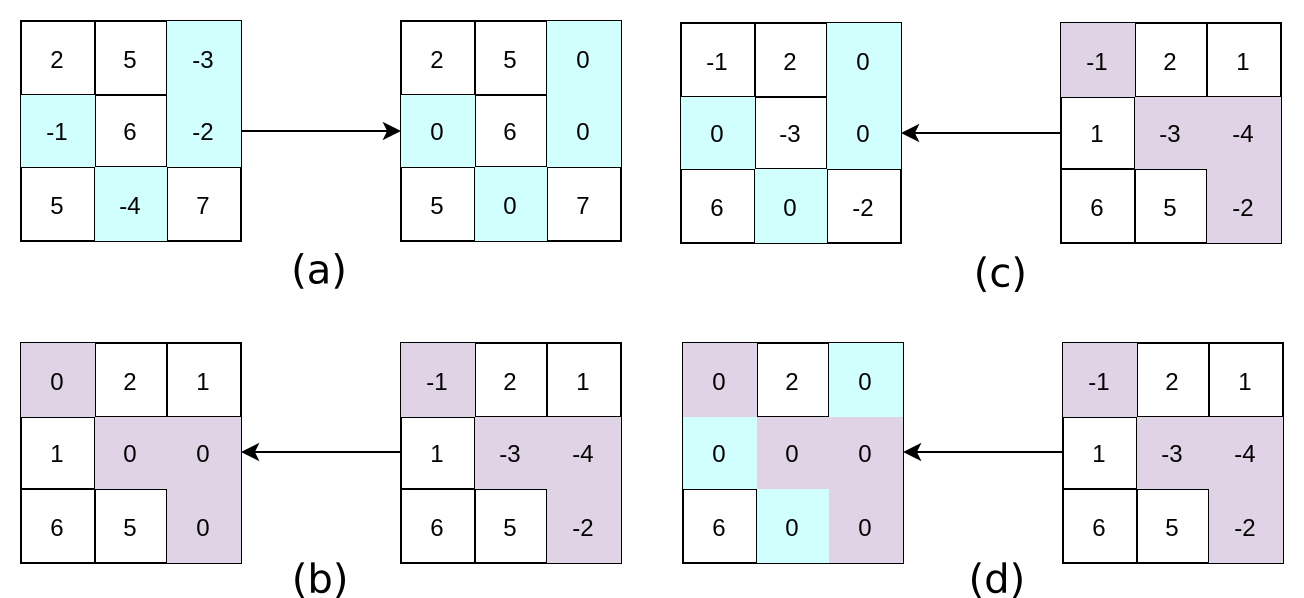}
    \caption{The ReLU operation during (a) Forward Pass (b) Backpropagation (c) Backpropagation with DeconvNet (d) Guided Backpropagation}
    \label{fig:guid_for_relu}
\end{figure}

\begin{figure}
    \centering
    \includegraphics[width=0.95\linewidth]{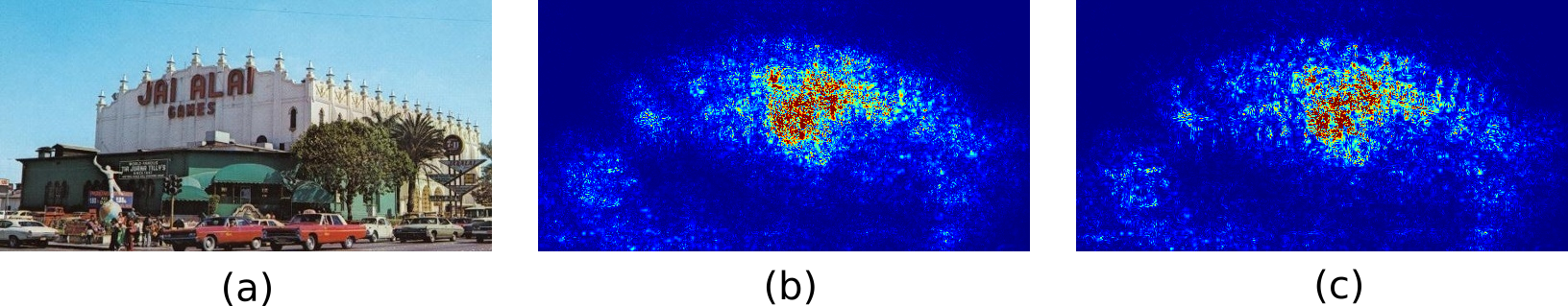}
    \caption{Samples showing the saliency maps for the (a) Sample image given by (b) Gradient backpropagation (c) Guided backpropagation methods. Sample taken from MexCulture Architectural Styles dataset\cite{obeso2016image}}
    \label{fig:grad_guided}
\end{figure}

Computing a saliency map based on gradients gives an idea of the various input features (pixels) that have contributed to the neuron responses in the output layer of the network. The primary idea proposed by Springenberg et al.\cite{springenberg2014striving} is to prevent the backpropagation of negative gradients found in the deconvolution approach as they decrease the activation of the higher layer unit we aim to visualize.
This is achieved by combining the rectification operation performed by the DeconvNet and the gradient backpropagation. As shown in Fig.~\ref{fig:guid_for_relu}(d), guided backpropagation proposes to restrict the flow of the gradients that have a negative value during backpropagation and also those values that had a negative value during the forward pass. This nullification of negative gradient values is called the \textit {guidance}. Using the guidance step results in sharper visualizations for the descriptive regions in the input image\cite{springenberg2014striving}.
Figure.~\ref{fig:grad_guided} shows the heat maps generated by the gradient backpropagation and guided backpropagation methods for the network trained on ResNet34 architecture\cite{montoya2019organizing}. It can be seen that the guidance steps result in reducing the number of pixels that have a higher importance score and hence produce slightly sharper visualization.


\subsection{Integrated Gradients}
\begin{figure}
    \centering
    \includegraphics[width=\textwidth]{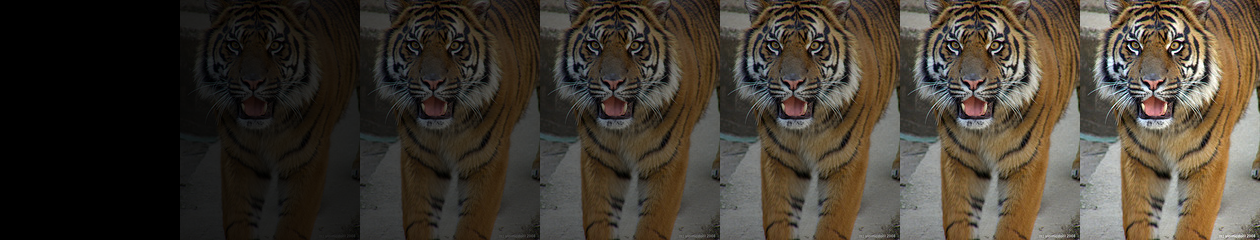}
    \caption{Transformation of the baseline image for integrated gradients for 7 values of $\alpha = [0,1]$. Image has been taken from ImageNet database\cite{imagenet_cvpr09}}
    \label{fig:int_grad}
\end{figure}

Sundararajan et al.\cite{sundararajan2017axiomatic} proposed to calculate the saliency map as an integration of the gradients for a set of images that are created from the transformation of a baseline image $x^{'}$ to the input image $x$. They propose the baseline $x^{'}$ as a black image, then the series of images is produced by a linear transformation $x(\alpha) = x^{'}+ \alpha \times (x-x^{'})$. Thus if we denote by $x_{i}$ the value of $i$-feature in our $x$ then Eq.~(\ref{eq:int_grad}) shows the calculation of the integrated gradients for the network with output classification for a class $c$ as $S_c(x)$. This forms the map of relevance scores $R^c$ with a corresponding score for each input pixel $x_i$. 

\begin{equation}
    R^c_{i} = (x_i - x_i^{'}) \times \int_{\alpha=0}^{1} \frac{\partial S_c(x^{'}+\alpha \times (x - x^{'})}{\partial x_i} d\alpha 
    \label{eq:int_grad}
\end{equation}
The parameter $\alpha$ varies in $[0,1]$ and the term within the partial derivative would go from the baseline image to the final input as we integrate over $\alpha$ as shown in Fig.~\ref{fig:int_grad}. In practice, the integral is approximated by a summation over a fixed number of samples i.e. Riemann approximation. 

The authors observed that if there were slight changes in the pixel value of the image $x$ such that visually the image did not appear to have changed, the gradients calculated by the gradient backpropagation methods showed large fluctuations in their values. For a small amount of noise present in the image, the visualization with gradient backpropagation was different to that of the original image. In the integrated gradients method an averaging is performed over a sample of images and so the final relevance maps would be less sensitive to these fluctuating gradient values when compared with the other gradient-based methods.

\subsection{SmoothGrad}
An alternative method to circumvent the issue of noisy saliency maps called the SmoothGrad was proposed by Smilkov et al.\cite{smilkov2017smoothgrad}. The idea of this method is to have a smoother map with sharper visualizations by averaging over multiple noisy maps. To achieve this, the authors propose to add a small noise sampled from Gaussian distribution $\mathcal{N}(0, \sigma^2)$, where $\sigma$ is the standard deviation, to the input image $x$ for each color channel. Thus, they create $n$ samples of the input image with a small amount of noise added to its pixels. The relevance score maps are calculated for each of these images and the average of these $n$ generated maps gives the final relevance score map for the image $x$ as shown in Eq.~(\ref{eq:smoothgrad}). SmoothGrad is not a standalone method rather it can be used as an extension to other gradient-based methods to reduce the visual noise of the saliency maps. The authors observe that adding about $10-20\%$ noise to sampled images produced sharper maps. The parameter $\sigma$ was chosen such that $\frac{\sigma}{x_{max} - x_{min}}$ was in the range of $[0.1, 0.2]$. Here, $x_{max}$ and $x_{min}$ refer to the maximum and minimum values of the pixels of the image.


\begin{equation}
\begin{split}
    x_i = x+ \mathcal{N}(0, \sigma^2)\\
    \hat{R^c}(x) = \frac{1}{n} \sum_{i=1}^n R^c (x_i)
\end{split}
    \label{eq:smoothgrad}
\end{equation}

\subsection{Gradient-Class Activation Mapping (Grad-CAM)}
\label{sec:Grad-CAM}

Gradient Class Activation Mapping (Grad-CAM)\cite{selvaraju2017grad} is a post-hoc explanation via visualization of class discriminative activations for a network. Similar to gradient-based methods, Grad-CAM leverages the structure of the CNN to produce a heat map of the pixels from the input image that contribute to the prediction of a particular class. 

A key observation that Grad-CAM relies on is that the deeper convolutional layers of a CNN act as high-level feature extractors\cite{bengio2013representation}. So the feature maps of the last convolution layer of the network would contain the structural spatial information of objects in the image. Therefore, instead of propagating the gradient till the input layer like other gradient-based methods, Grad-CAM propagates the value from the output till the last convolutional layer of the network.

The features maps from the last convolution layer cannot be used directly as they would contain information regarding all the classes present in the dataset. Assuming that the last convolution layer of the network has $k$ feature maps named $A^1, A^2, ..., A^k$, the Grad-CAM method proposed to determine an importance value for each of the $k$ maps to the class $c$ predicted by the network. This value is calculated as the global average pooling of the gradient of the classification score $S_c(x)$ with respect to the activation values $A^k$ for that feature map. As shown in Eq.~(\ref{eq:grad_weights}), $\alpha_k^c$ is the importance value for the feature map $k$ and there are $k$ such weights that are computed. 

\begin{equation}
    \alpha_k^c = \frac{1}{Z}\sum_i \sum_j \frac{\partial S_c(x)}{\partial A^k_{ij}}  
    \label{eq:grad_weights}
\end{equation}

Here $Z = h \times w$ where $h$ and $w$ correspond to the height and width dimension of each feature map.

The $\alpha_k^c$ weights are then used to weight each feature in the $k$ feature maps, the latter are then averaged. This gives us the relevance score map $R^c$ and a ReLU (rectification) function is applied over this map, see Eq.~(\ref{eq:gradcam}), to nullify the features that are negative and retain only those values that have a positive influence. At this stage, the relevance map $R^c$ is a 2-D map with the same spatial dimension as the feature maps of the last convolution layer. To have a correspondence to the input image $x$, $R^c$ is upsampled to the spatial dimension of $x$ using interpolation methods and scaled to the interval of $[0,1]$ to visualize the final heat map.


\begin{equation}
    R^c = ReLU \left ( \sum_k \alpha^c_k A^k \right )
    \label{eq:gradcam}
\end{equation}


Grad-CAM is the generalization of previously proposed by Zhu et al.\cite{zhou2016learning} CAM method which requires squeezing of the feature maps of the last conv layer by average pooling to form the input of the FC layer in the network. Grad-CAM on the contrary can be applied to all the architectures of Deep CNNs.

\subsubsection{Guided Grad-CAM}
\begin{figure}
    \centering
    \includegraphics[width=0.95\linewidth]{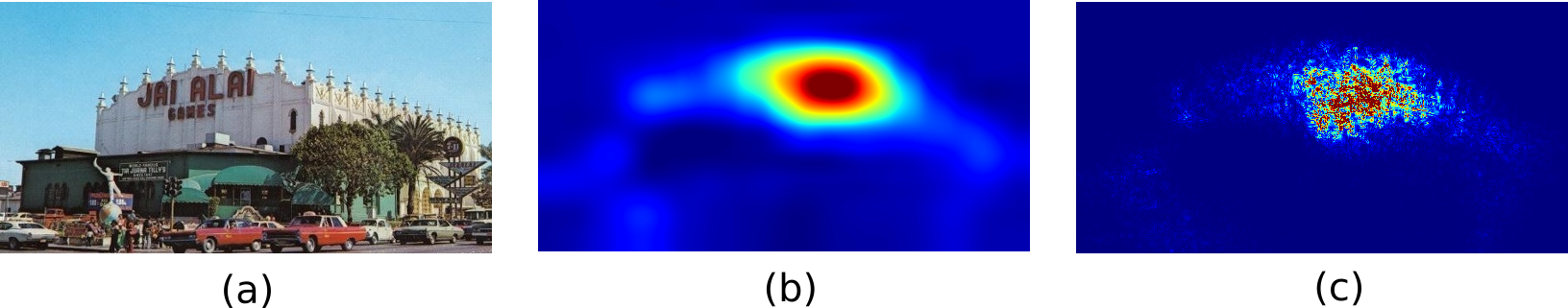}
    \caption{Samples showing the saliency maps for the (a) Sample image (b) Grad-CAM (c) Guided Grad-CAM methods. Image taken from MexCulture Architectural Styles dataset\cite{obeso2016image}}
    \label{fig:grad_cam_guided}
\end{figure}

The heat maps produced by Grad-CAM are coarse, unlike the other gradient-based methods\cite{selvaraju2017grad}. As the feature map of the last convolutional layer has a smaller resolution compared to the input image $x$, Grad-CAM maps do not have fine-grained details that are generally seen in other gradient-based methods. To refine the maps, a variant of the method called the Guided Grad-CAM has been proposed which is a combination of the Grad-CAM and the guided backpropagation method by doing an element-wise multiplication of the two maps. The heat map that is obtained by this operation has been observed to have a higher resolution\cite{selvaraju2017grad}. We illustrate maps obtained by Grad-CAM and Guided Grad-CAM in Fig.~\ref{fig:grad_cam_guided}.

\section{Methods based on network structure} \label{sec:net_arch}
This category of methods integrates the architecture of the network while explaining the output. Starting from an output neuron, they employ different local redistribution rules to propagate the prediction to the input layer to obtain the relevance score maps. Unlike the previous category, these methods do not compute gradients in the network. In this section, we present the details of them.

\subsection{Layer-wise Relevance Propagation (LRP)} \label{sec:lrp}
Layer-wise Relevance Propagation (LRP) is an explanation method proposed by Bach et al.\cite{bach2015pixel} that explains the decision of a network for a particular image by redistributing the classification score $S_c$ for a class $c$ backwards through the network. The method does not use gradient calculations but defines the activation of the output neuron (either the predicted class or another class that is being considered) as the relevance value and a set of local rules for the re-distribution of this relevance score backwards till the input, layer by layer. The first rule that they propose is that of \textit{relevance conservation}. Let the neurons in the different layers of the network be denoted by $\nu, \xi, o$ etc. and $S_c(x)$ the classification score for the input image $x$ regarding the class $c$. Then according to the relevance conservation rule, the sum of the relevance scores of all the neurons in each layer is a constant and equals $S_c(x)$ as shown in Eq.~(\ref{eq:rel_cons}).


\begin{equation}
    \sum_\nu R_\nu = . . . = \sum_\xi R_\xi = \sum_o R_o = ... = S_c(x)
    \label{eq:rel_cons}
\end{equation}


Let $l$ and $l+1$ be two consecutive layers in the network and $i$, $j$ denote the neurons belonging to these layers respectively. The relevance of the neuron $j$ based on $S_c(x)$ can be written as $R^c_j$. If neuron $i$ is connected to neuron $j$ then it is assigned a relevance value of $R^c_j$ weighted by the activation of the neuron $a_i$ and the weight of the connection between the two neurons $w_{ij}$. Similarly, neuron $i$ receives a relevance value from all the neurons that it is connected to in the next layer $(l+1)$. The sum of all the relevance contributions that the neuron receives from the neurons it was connected to in the next layer is the final relevance value $R_i^c$ that is assigned to the neuron as shown in Eq.~(\ref{eq:lrp}). The denominator term in Eq.~(\ref{eq:lrp}) is the normalization value that is used to ensure relevance conservation rule Eq.~(\ref{eq:rel_cons}). This rule is termed as the LRP-0 rule\cite{montavon2019layer}. 



\begin{equation}
    R^c_{i} = \sum_j \frac{a_i w_{ij}}{\sum_{0,q} a_q w_{qj}} R^c_{j}
    \label{eq:lrp}
\end{equation}

In this equation, the summation $\sum_{0,q}a_i w_{qj}$ is done for all the neurons in the lower layer $q = 0, ..., Q_l$ including the bias neuron in the network. The activation of the bias neuron is considered as $a_0 = 1$ and the weight of the connection is denoted as $w_{0j}$. For the relevance propagation, the bias neuron is considered only for this term and is not considered elsewhere. 
Note that the authors propose these rules only for the specific case of rectifier networks i.e. for networks with ReLU as the non-linearity. The relevance of the output neuron is considered as its activation taken before the Softmax layer.  

Similarly there exist a few other rules that improve on the LRP-0 rule for the propagation of relevance as presented in the following list.

\begin{itemize} 
    \item Epsilon Rule (LRP-$\epsilon$): To improve the stability of the LRP-0 rule a small positive term $\epsilon$ is added to the denominator as shown in Eq.~(\ref{eq:lrp_eps}). The $\epsilon$ term also reduces the flow of the relevance if the activation of the neuron is very small or there is a weak connection between the two neurons. If the value of $\epsilon$ is increased then it aids in ensuring only the stronger connections receive the redistributed relevance\cite{montavon2019layer}.
    \begin{equation}
    R^c_{i} = \sum_j \frac{a_i w_{ij}}{\sum_{0,q} a_q w_{qj} + \epsilon} R^c_{j}
    \label{eq:lrp_eps}
\end{equation}
    \item LRP-$\gamma$ : The parameter $\gamma$ was introduced to improve the contributions of the connections that had a positive weight ($w_{ij}^+$) as shown in Eq.~(\ref{eq:lrp_gamma}). The function $(f)^+ = \max (0, f)$ and so the neurons with a positive weight connection receive a higher relevance score during propagation. 
     \begin{equation}
    R^c_{i} = \sum_j \frac{a_i \cdot (w_{ij} + \gamma w_{ij}^+)}{\sum_{0,q} a_q \cdot (w_{qj} + \gamma w_{qj}^+)} R^c_{j}
    \label{eq:lrp_alpha_beta}
\end{equation}
    \item LRP $\alpha\beta$ rule: Two parameters $\alpha$ and $\beta$ are used to control separately the positive and negative contributions to the relevance propagation. The function $(f)^+ = \max(0,f)$ and $(f)^- = \min(0,f)$ and the parameters are constrained under the rule of $\alpha = \beta +1$.
    
     \begin{equation}
    R^c_{i} = \sum_j \left ( \alpha \frac{(a_i w_{ij})^+}{\sum_{0,q}(a_q w_{qj})^+}  - \beta \frac{(a_i w_{ij})^-}{\sum_{0,q}(a_q w_{qj})^-} \right ) R^c_{j}
    \label{eq:lrp_gamma}
\end{equation}
\end{itemize}



\subsubsection{LRP as Deep Taylor Decomposition (DTD)}
Montavon et al.\cite{montavon2017explaining} propose a framework to connect a rule-based method like LRP and the Taylor decomposition method as a way to theoretically explain the choice of the relevance propagation rules\cite{bach2015pixel}. They propose a method called the Deep Taylor Decomposition (DTD) which treats LRP as consecutive Taylor expansions applied locally at each layer and neuron. The main idea that DTD uses is that a deep network can be written as a set of subfunctions that relate the neurons of two consecutive layers. Instead of treating the whole network as a function $f$, DTD expresses LRP as a series of mapping of neurons $i$ at a layer $l$ to the relevance $R_j$ of the neuron $j$ in layer $l+1$. 

The Taylor expansion of the relevance score $R_j$ can be expressed as function of the activations $a_i$ 
at some \textit{root} point $\tilde{\textbf{a}}$ in the space of the activations as shown in Eq.~(\ref{eq:taylor_exp_rel}).

\begin{equation}
    R_j(\textbf{a}) = R_j(\tilde{\textbf{a}}) + \sum_{0,i} (a_i - \tilde{a_i}) [\nabla R_j (\tilde{\textbf{a}})]_i + ...
    \label{eq:taylor_exp_rel}
\end{equation}

The first-order term in this expansion can be used to determine how much of the relevance $R_j$ is redistributed to the neurons in the lower layer. The main challenge in the computation of the Taylor expansion, in this case, is that of finding the appropriate \textit{root} point $\tilde{\textbf{a}}$ and compute the local gradients. 

To compute the function $R_j(\textbf{a})$, the authors propose to substitute it with a relevance model that is simpler to analyze. From the relevance propagation rules of LRP (Sec.~\ref{sec:lrp}), it can be seen that the relevance score of a neuron can be written as function of its activations as $R_i = a_i \cdot r_i$, where $r_i$ in the case of the LRP-0 rule Eq.~(\ref{eq:lrp}) can be written as $r_i = \sum_i \frac{w_{ij}}{\sum_{0,i} a_i w_{ij}} R_j$. As the LRP rules are described for deep rectifier networks, the relevance function $R_j(\textbf{a})$ is expressed based on the ReLU activation as:
\begin{equation}
    R_j(\textbf{a}) = \max(0, \sum_{0,i} a_i w_{ij}) \cdot r_j
\end{equation}

A Taylor expansion of this function gives:
\begin{equation}
    R_j(\textbf{a}) = R_j(\tilde{\textbf{a}}) + \sum_{0,i} (a_i - \tilde{a_i}) \cdot w_{ij} r_j 
    \label{eq:dtd_Final}
\end{equation}

Due to the linearity of the ReLU function on the domain of positive activations the higher order terms in the expansion are zero. The choice of the \textit{root} point would ensure that the zero order terms can be made small. The first order term computation is fairly straightforward and would identify how much of the relevance value $R_j$ should be redistributed to the neurons of the lower layer. Different LRP rules that have been presented previously can be derived from Eq.~(\ref{eq:dtd_Final}) based on the choice of the reference point $\tilde{\textbf{a}}$. For instance, LRP-0 rule shown in Eq.~(\ref{eq:lrp}) can be derived by choosing $\tilde{\textbf{a}} = 0$ and LRP-$\epsilon$ as shown in Eq.~(\ref{eq:lrp_eps}) by choosing $\tilde{\textbf{a}} = \epsilon \cdot (a_j + \epsilon)^{-1}$.



\subsection{Deep Learning Important FeaTures (DeepLIFT)} \label{sec:deeplift}

The primary idea of DeepLIFT a method proposed by Shrikumar et al.\cite{shrikumar2017learning} is similar to the LRP method explained in Sec.~\ref{sec:lrp}. The major difference between the two methods is that DeepLIFT establishes the importance of neurons in each layer in terms of the difference of their response to that one of a \textit{reference state}. The reference state is either a default image or is an image chosen based on domain-specific knowledge. This reference could be an image that has the specific property against whose differences in the explanations are meant to be calculated. For example, it could be a black image in the case of the MNIST dataset as the backgrounds of the images in that dataset are all black. DeepLIFT aims to explain the difference in the output produced by the input image and the output of the reference state based on the difference between the input image and the chosen reference image.

For the output classification score $S_c$ of the input image $x$ and the output score of reference state as $S_c^0$, the difference term $\Delta S_c$ is defined as $S_c - S_c^0$. For the neurons $y_1, ...,y_i, ..., y_n$ belonging to a layer the relevance is denoted by $R_{\Delta y_i , \Delta S_c}$. $\Delta y_i$ denotes the difference between the activations of the neuron $y_i$ for the input image and the reference state. Similar to the LRP method, DeepLIFT has the \textit{Summation to delta} rule where the summation of the relevance of neurons at each is constant and equal to $\Delta S_c$ as shown in Eq.~(\ref{eq:deeplift}). 

\begin{equation}
    \sum_{i=1}^{n} R_{\Delta y_i, \Delta S_c} = \Delta S_c
    \label{eq:deeplift}
\end{equation}


In order to explain the propagation rules, the authors define the a term \textit{multiplier} - $m_{\Delta y \Delta S_c}$, which is defined as the contribution $R_{\Delta y, \Delta S_c}$ of the difference in the reference and input image activations $\Delta y$ to the difference in the output prediction $\Delta S_c$ divided by $\Delta y$ as shown in Eq.~(\ref{eq:multiplier}).

\begin{equation}
    m_{\Delta y \Delta S_c} = \frac{R_{\Delta y \Delta S_c} }{ \Delta y}
    \label{eq:multiplier}
\end{equation}

The multiplier is a term that is similar to a partial derivative but defined over finite differences\cite{shrikumar2017learning}. They also define the \textit{Chain rule for multipliers} similar to the chain rule used with derivatives as shown in Eq.~(\ref{eq:chain_rule}) where $z_j$ denotes the neurons in intermediate layers between the neurons $y$ and the output $S_c$.

\begin{equation}
    m_{\Delta y_i \Delta S_c} = \sum_j m_{\Delta y_i \Delta z_j}m_{\Delta z_j \Delta S_c}
    \label{eq:chain_rule}
\end{equation}

Similar to LRP, the authors also separate the relevance values into two terms: positive and negative as they can then be treated differently if required. For each neuron $y$, the two terms $\Delta y^+$ and $\Delta y^-$ are the positive and negative components respectively. These components can be found by grouping the positive and negative terms that contribute to the calculation of $\Delta y$. Based on this idea the difference in the neuron activations of the input image and the reference state, and the relevance contribution can be written as shown in Eq.~(\ref{eq:pos_neg}).

\begin{equation}
\centering
\begin{split}
    \Delta y = \Delta y^+ + \Delta y^-\\
    R_{\Delta y \Delta S_c} = R_{\Delta y^+ \Delta S_c} + R_{\Delta y^- \Delta S_c} 
\end{split}  
\label{eq:pos_neg}
\end{equation}

Using these terms, DeepLIFT proposes three rules that can be applied to a network for different layers to propagate the relevance from the output to the input layer.

\begin{itemize}
    \item \textbf{Linear Rule}: The linear rule is applied for the FC and Convolution layers (not applicable for the non-linearity layers). Considering the function $z =  \sum_i w_i y_i + b$, where $z$ is the activation of the neuron in the next layer, $y_i$ are the activations to the neuron from the previous layer and $w_i$ is weights of the connections, then we have $\Delta z = \sum_i w_i \Delta y_i$ based on the difference taken with the activations of the reference state neurons. The relevance contribution is then written as $R_{\Delta y_i \Delta z} = w_i \Delta y_i$. The multiplier in this case is given by $m_{\Delta y_i \Delta z} = w_i$.

    \item \textbf{Rescale Rule}: The Rescale rule is applied to layers with the non-linearities like the ReLU. Consider the neuron $z$ to be the non linear transformation of $y$ as $z = g(y)$. In the case of the ReLU function (Eq.~(\ref{eq:relu})) denoted by $g(y)$. Considering the summation to delta property, the relevance contribution is $R_{\Delta y \Delta z}  = \Delta z$ as there is only one input $y$. Hence, the multiplier in this case would be $m_{\Delta y \Delta z} = \frac{\Delta z }{\Delta y}$. 
    
    \item \textbf{RevealCancel Rule}: The RevealCancel rule treats the positive and negative contributions to relevance values separately. The impact of the positive and negative components of $y$ given as $\Delta y^+$ and $\Delta y^-$ on the components of $z$ given by $\Delta z^+$ and $\Delta z^-$ are calculated separately. Instead of the straightforward calculation, the value of $\Delta z^+$ is computed as the average of two terms. The first term is the average impact of the addition of only $\Delta y^+$ terms on the output of the nonlinearity $g$. With $y^0$ as the value of the reference state at that neuron, the impact of $\Delta y^+$ is calculated by comparing the difference in the function value when it is included on top of $y^0$, and is given as $g(y^0 + \Delta y^+) - g(y^0)$. 
    The second term computes the impact of $\Delta y^+$ after the negative terms have been included. So the term $g(y^0 + \Delta y^+ +  \Delta y^-) - g(y^0 + \Delta y^-)$ measures the impact of $\Delta y^+$ when both the reference and negative terms are present. This computation is shown in Eq.~(\ref{eq:rvcnl_1}). 
    \begin{equation}
        \Delta z^+ = \frac{1}{2} (g(y^0 + \Delta y^+) - g(y^0)) + \frac{1}{2} (g(y^0 + \Delta y^+ +  \Delta y^-) - g(y^0 + \Delta y^-))
         \label{eq:rvcnl_1}
    \end{equation}    
    
    Similarly for the calculation of $\Delta z^-$, the average individual $\Delta y^-$ term is first computed in the absence of the positive term $\Delta y^+$ and then another term with the inclusion of $\Delta y^+$ is added to get the total impact as shown in Eq.~(\ref{eq:rvcnl_2}).
        \begin{equation}
        \Delta z^- = \frac{1}{2} (g(y^0 + \Delta y^-) - g(y^0)) + \frac{1}{2} (g(y^0 + \Delta y^- +  \Delta y^+) - g(y^0 + \Delta y^+))
         \label{eq:rvcnl_2}
    \end{equation}    
    
    Thus two multipliers that will be computed using this rule are as shown in Eq.~(\ref{eq:rvcnl_3}), where $\Delta z^+$ and $\Delta z^-$ are calculated using Eqs.~(\ref{eq:rvcnl_1}), (\ref{eq:rvcnl_2}) and $\Delta y^+$ and $\Delta y^-$ correspond to the sum of the positive and negative terms of $\Delta y$.
    \begin{equation}
        \begin{split}
            m_{\Delta y^+ \Delta z^+ } = \frac{R_{\Delta y^+ \Delta z^+ } }{\Delta y^+} = \frac{\Delta z^+}{\Delta y^+}\\
        m_{\Delta y^- \Delta z^-} = \frac{R_{\Delta y^- \Delta z^- } }{\Delta y^-} = \frac{\Delta z^-}{\Delta y^-}
        \end{split}
        \label{eq:rvcnl_3}
    \end{equation}
    The authors propose that the relevance scores that have been assigned to $\Delta y^+$ and $\Delta y^-$ are then distributed to the input features using the Linear Rule.
    
\end{itemize}

\subsection{Feature based Explanation Method (FEM)}
Feature based Explanation Method (FEM) proposed by Fuad et al.\cite{fuad2020features}, similar to Grad-CAM, employs the observation that deeper convolutional layers of the network act as high-level feature extractors. 

Let us consider a CNN that comprises a single Gaussian filter at each convolution layer. Then the consecutive convolutions of the input image $x$ with the Gaussian filters followed by the pooling (downsampling) would be the same operation that would be performed to create a multi-resolution Gaussian pyramid. In this pyramid, the image at the last level would have only the spatial information of the main objects which are present in the image. Considering a standard CNN, the learned filters at the deeper convolution layers behave similar to the high-pass, i.e. derivative filters on the top of Gaussian pyramid (some examples are given in\cite{zemmari2020deep}). This would imply that the information contained in the feature maps of the last convolution layer correspond to the main object that has been detected by the network from the given input image $x$.

Hence, FEM proposes that the contribution of the input pixels for the network decision can be directly inferred from the features of $x$ that have been detected at the last conv layer of the network. Also, FEM proposes that the final decision of the output would be influenced by the \textit{strong} features from $k$ maps in the last conv layer. FEM supposes that the $k$ feature maps of the last convolutional layer have a Gaussian distribution. In this case, the \textit{strong} features from these maps would correspond to the \textit{rare} features. The authors propose a K-sigma filtering rule to identify these \textit{rare} and \textit{strong} features. Each of the feature maps in layer $k$ is thresholded based on the K-sigma rule to create binary maps $B_k(a_{i,j,k})$ where $i,j$ denote the spatial dimension of the $k$ feature maps as shown in Eq.~(\ref{eq1}). The mean $\mu_k$ and standard deviation $\sigma_k$ are calculated for each of the $k$ feature maps followed by the thresholding to create $k$ binary maps $(B_k)$. K is the parameter that controls the threshold value and is set to 1 by the authors in their work.
\begin{equation}
  B_k(a_{i,j,k}) = 
     \begin{cases}
       \text{1} &\quad\text{if $a_{i,j,k} \geq \mu_k + K * \sigma_k$}\\
       \text{0} &\quad\text{otherwise}
     \end{cases}
     \label{eq1}
\end{equation}

The hyperparameters of a DNN such as the number of filters to train at each layer is often set arbitrarily, as the hyperparameter optimization process is a heavy computational problem. Hence channel attention mechanisms have been proposed in the DNNs\cite{hu2018squeeze} to improve classification accuracy. These mechanisms select important feature channels (maps) in an end-to-end-training process. Inspired by these models, the authors hypothesize that not all feature maps will be important for the classification. The importance is always understood as the magnitude of positive features in the channels. Hence, a weight term equal to the mean of the initial feature maps $w_k = \mu_k$ is assigned to each of the binary maps $B_k$. The importance map $R$ is computed as the linear combination of all the weighted binary maps $B_k$ and then normalized to the interval $[0,1]$. The importance map $R$ is upsampled to the same spatial resolution as the input image $x$ by interpolation.

\begin{figure}
    \centering
    \includegraphics[width=\linewidth]{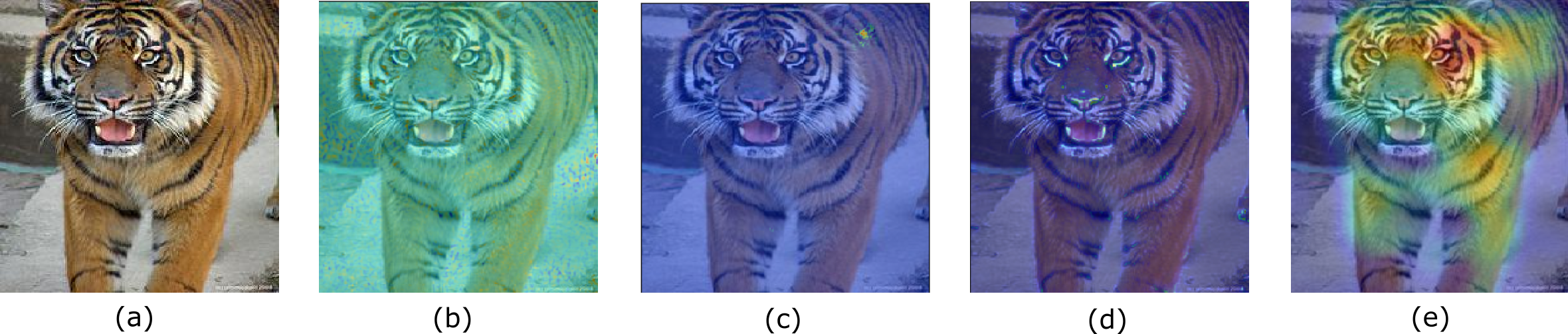}
    \caption{Explanation heat maps obtained for (a) Sample Image by (b) LRP-$\epsilon$ (c) LRP-$\epsilon$ ignore bias (d) LRP-$\alpha\beta$ with $\alpha=1$, $\beta=0$ (e) FEM}
    \label{fig:fem_lrp}
\end{figure}

FEM eliminates the need to calculate the gradients from the output neuron and provides a faster and simpler method to get an importance score of the input pixels based only on the features that have been extracted by the network. It does not examine the classification part of the network but uses only the feature extraction part of the CNN to explain the important input pixels that have been extracted by the network to produce the decision.
The method is applicable both for 2D and 3D images or video, considered as a 2D +t volume. We will now illustrate it in the problem of image classification from ImageNet database performed with the VGG16\cite{SimonyanZ14a}. We propose the reader to visually compare the heat maps presented in Fig.~\ref{fig:fem_lrp} obtained by using different LRP rules\cite{bach2015pixel} and FEM. 

It can be seen from the fgure that LRP heat maps are dependent on the rule that is used. In this case, the LRP-$\epsilon$ assigns equal weight to both positive and negative features, and in the Fig.~\ref{fig:fem_lrp}(a) most of the input pixels get assigned a higher relevance score.
LRP-$\epsilon$ without bias from Fig.~\ref{fig:fem_lrp}(c) results in a heat map that has the importance only in a smaller region of the tiger near the top. Without the added bias term the relevance scores are not distributed properly to the other regions. 
The $\alpha\beta$ rule with $\alpha=1$ and $\beta=0$ only considers the features with a positive influence. As illustrated in the figure, this rule only highlights the contours with higher contrast in the image.
Though FEM also considers the positive features (features are taken after the ReLU), the heat map that is more holistic and highlights the important regions in the image.

\section{Methods based on Adversarial approach} \label{sec:adversary}
The adversarial approach for explanations is more recent and differs in its principles from previously presented categories. They are taken from the principles of adversarial attacks and the use of generative networks like Generative Adversarial Networks (GANs) to explain the DNN classifiers. Thus adversarial methods deserve a separate category.

Many recent works have used adversarial attacks on CNNs to demonstrate the susceptibility of the networks to simple methods that could lead the network to make completely wrong predictions\cite{szegedy2013intriguing}. Different adversarial attacks on the network can be used to interpret its behaviour\cite{tao2018attacks}. Sample images that produce adversarial results give hints about the behaviour of the network\cite{ross2018improving, IgnatievNM19}. For example, the one-pixel attack proposed by Su et al.\cite{su2019one}, showed that the network can have a completely wrong prediction when just one pixel in input image has been changed. Studying these adversarial attacks we can interpret the regions of the image that the network focuses on to make a decision.

In addition to the adversarial attack based interpretation of the network, we bring focus on a recent adversarial learning-based explanation method proposed by Charachon et al.\cite{charachon2020combining} that uses a Generative Adversarial Network (GAN)\cite{GoodfellowPMXWOCB14} based model. GAN is a type of network architecture with two components: i) Generator ii) Discriminator, that simultaneously trains the generator (G) to learn the data distribution and the discriminator (D) to estimate if the sample belongs to the dataset or has been generated by G. 

For the case of a binary classification task on medical images, the authors \cite{charachon2020combining} along with their CNN classifier, use two generator networks i) Similar image generator $\bar{g_s}$ ii) Adversarial image generator $\bar{g_a}$ to produce explanations. The network $\bar{g_s}$ is trained to generate an image that has the same output as the input image $x$ by the network. The $\bar{g_a}$ network is trained to produce an image with a prediction that is opposite to that of the $x$ (adversarial). Consequently, the authors propose that the difference in these two generated images forms the explanation of the output by the network. For a given image $x$ the explanation of the classifier is then given as shown in Eq.~(\ref{eq:adver}).
\begin{equation}
    R(x) = |\bar{g_s} (x) - \bar{g_a}(x)|
    \label{eq:adver}
\end{equation}
The approach of just one network to generate an adversarial image ($x_a)$ and use the difference in the input image $x$ and the $x_a$ as the explanation was observed to produce noisy and non-intuitive features. 
To improve this, the authors use the two-generator approach and train $\bar{g_s}$ and $\bar{g_a}$ to sample from the same adversarial space, to have minimal differences in their learnt parameters but produce images with opposite classifications by the CNN.

\section{Evaluation of explanation methods} 
\label{sec:comparison}
The evaluation of explanation methods remains an open research question. Today we can observe two trends in the evaluation of explanation methods. The first one is perturbation based when the input, dataset or network is perturbed and the induced variation in the importance maps serves as an explainer. The second one is an attempt to correlate the explainer with the human understanding of visual scenes and assess it with user feedback. For comparison of explanation maps, usual metrics for saliency maps are computed \cite{bylinskii2018different} such as Spearman Correlation Coefficient (SCC), Pearson Correlation Coefficient (PCC), Structural Similarity Index Measure (SSIM) and L1 similarity to name a few.

\subsection{Perturbation based evaluation}

Samek et al.\cite{samek2017evaluating} argue that heatmaps are representative of a classifier's view and will not necessarily have to follow human intuition or highlight the primary object in the image. Hence, to assess the relevance of the heat maps they order the pixels in increasing importance score associated with then and then iteratively perturb the pixels in the neighbourhood of these pixels. They then compute the mean difference of classification scores on the whole test dataset each time a perturbation is applied. If the mean difference of scores for a given class is high then the explanation heat-map is relevant, i.e. perturbing a region assigned a high importance score changes the output score of the network. For this, they compute the Area Over Perturbation Curve (AOPC)  metric and show that the LRP method (Sec.~\ref{sec:lrp}) is the best when compared to Gradient Backpropagation (Sec.~\ref{sec:grad_backprop}) and Deconvolution (Sec.~\ref{sec:deconv_net}) accordingly to the perturbation approach.

Furthermore, methods based on the computation of gradients like the backpropagation, guided backpropagation and DeconvNet suffer from gradient shattering\cite{balduzzi2017shattered}, where, as the depth of the layers increases the gradients of the loss progressively resembled white noise. This causes the importance values to have high-frequency variations and makes them highly sensitive to small variations in the input. Galli et al.\cite{galli2020} perform adversarial perturbations to their input image using the fast gradient method \cite{GoodfellowSS14} and DeepFool\cite{moosavi2016deepfool}. They compare the explanation maps of Guided Grad-CAM for the image and its perturbed variations using the Dice similarity coefficient after thresholding of importance map with 0.9 as the threshold value. They observe that perturbations in the image strongly affected the saliency maps. The maps of the images before and after perturbations are different, though they note that these differences are not easily perceived by a human viewer. LRP, DeepLIFT and FEM are not sensitive to this problem as they do not compute gradients.

Adebayo et al.\cite{Adebayo2018} recently proposed two randomization tests to assess the quality of explanation methods. The first one is the model parameter randomization. It consists of comparing the explanations given by a method for a randomly initialized untrained network and a trained network with the same architecture. The model parameters like the weights are randomized for the whole network and also layer by layer. The second test is data randomization, which consists of label randomization: on the contrary to \cite{samek2017evaluating}, the input data remain unchanged, but the labels of classes are randomly switched. In both cases, if the explainer is good, the resulting explanation maps will strongly differ from the explanation maps of correctly trained classifiers. As comparison metrics, they have used SSIM, SCC and PCC of histogram of gradients of explanation maps. The lower value of the metrics, the higher is the difference in explanation maps for these tests and hence better is the explainer. Comparing six white-box methods from the linearization category, see Sec.~\ref{sec:linear}, they conclude that the Grad-CAM and Gradient Backpropagation pass this test and are better at explaining the network from the data and parameters that have been learnt.

\subsection{Human Perception based evaluation}

The perturbation-based methods give a way to assess the quality of the explainer without any reference to human perception. A desirable property of every explanation map is that it is non-random and highlights only the relevant regions in the image and no more. Nevertheless, the question arises on what is the target of explanations. In most cases the target is a human and they are often not AI experts. Therefore, many methods use a qualitative assessment based on human inspection to evaluate and compare different maps. To evaluate the Grad-CAM maps, the authors of \cite{selvaraju2017grad} conduct user surveys to find which maps and models they find reliable.  
Cruciani et al.\cite{cruciani2020} demonstrate the usefulness of LRP to visualize relevant features on brain magnetic resonance imaging modality (MRI) for multiple sclerosis classification. It is observed that the LRP heat maps are sparse and not always as intuitive which is also illustrated by Fig.~\ref{fig:fem_lrp}. It is therefore important to consider the user, who will be using the maps while choosing a method to explain a network. For the domain of medical images, the specialist might require a map that provides holistic explanations for them to be able to interpret and trust the network decision. Although human measurements on the quality of these maps are useful, they are time-consuming, could introduce bias and inaccurate evaluations\cite{buccinca2020proxy}.

Another way to evaluate explainers consists of comparing the explanation maps with human understanding of visual content. The objective way to perform this is by calculating the comparison metrics between explanation maps and Gaze Fixation Density Maps (GFDMs) of human experts performing the same visual classification task as the trained network. This can also be done by a comparison of explanation maps and important regions in the images identified by human experts by manual contouring. Mohseni et al.\cite{MohseniBR21} perform a similar experiment. The final goal of explanations is to increase a user's trust in AI systems and especially for image classification tasks in our case. Accordingly, Mohseni et al.\cite{MohseniBR21} observe that providing nonsensical explanations (i.e., those that do not align with users’ expectations) may harm the users’ reported trust and observed reliance on the system.

\begin{table}
\centering
\begin{tabular}{c|c|c}
\hline
\textbf{Method}                                                    & \textbf{Sim}            & \textbf{PCC}            \\ \hline
\begin{tabular}[c]{@{}c@{}}Gradient\\ Backpropagation\end{tabular} & 0.4982 $\pm$ 0.051          & 0.2422 $\pm$ 0.142          \\
\begin{tabular}[c]{@{}c@{}}Guided \\ Backpropagation\end{tabular}  & 0.5724 $\pm$ 0.063           & 0.0779 $\pm$ 0.067          \\
SmoothGrad                                                         & 0.5769 $\pm$ 0.062           & 0.1867 $\pm$ 0.113           \\
\begin{tabular}[c]{@{}c@{}}Integrated \\ Gradients\end{tabular}    & 0.5747 $\pm$ 0.063            & 0.1390 $\pm$ 0.082          \\
Grad-CAM                                                           & 0.5857 $\pm$ 0.104          & \textbf{0.3969 $\pm$ 0.290} \\
LRP                                                                & 0.5728 $\pm$ 0.063          & 0.0754 $\pm$ 0.069          \\
DeepLIFT                                                               & 0.5846  $\pm$ 0.063          & 0.2749 $\pm$ 0.081          \\
FEM                                                                & \textbf{0.5999 $\pm$ 0.087} & 0.3558 $\pm$ 0.289           \\ \hline
\end{tabular}
\caption{Comparison of explanation maps of different methods with the Gaze Fixation Density Maps (GFDM) for the VGG16 network trained on the MexCulture dataset\cite{DBLP:conf/ipta/ObesoBGGGR18} }
\label{tab:res}
\end{table}

Thus, in our experiments for an initial evaluation of the methods presented in this paper, we follow the strategy of comparison of explanation maps with user expectations for a given visual classification task. We measure them by GFDMs built upon gaze fixations of observers in a task-driven psychovisual experiment. The usual metrics of comparison of saliency maps, such as PCC, SSIM, L1 norm of difference (similarity) can also be used here. The higher the value of correlation, the stronger is the trust of the user in the explainer. Hence in this paper, we compare the presented white-box methods on the datasets with human gaze fixations recorded in a task-driven visual experiment of recognition of architectural styles of Mexican architectural heritage, introduced in \cite{DBLP:conf/ipta/ObesoBGGGR18} and publicly available at \footnote{https://api.nakala.fr/data/11280\%2F5712e468/1e412e0a43b5635365293b249feb9d53d74b5dc8}.
As comparison metrics, we have chosen PCC and  L1 norm of difference (SIM). We have taken a VGG-16 network initialized with ImageNet weights and retrained it on the complete Mexculture dataset from \cite{ObesoBGR19}. The results of the comparison of explanation maps from different white-box methods and the GFDM for this network are shown in Tab.~\ref{tab:res}. Here the metrics have been computed on 284 images of the MexCulture dataset with available gaze fixation density maps for images that have not been used for training.

\begin{figure}
    \centering
    \includegraphics[width=0.9\linewidth]{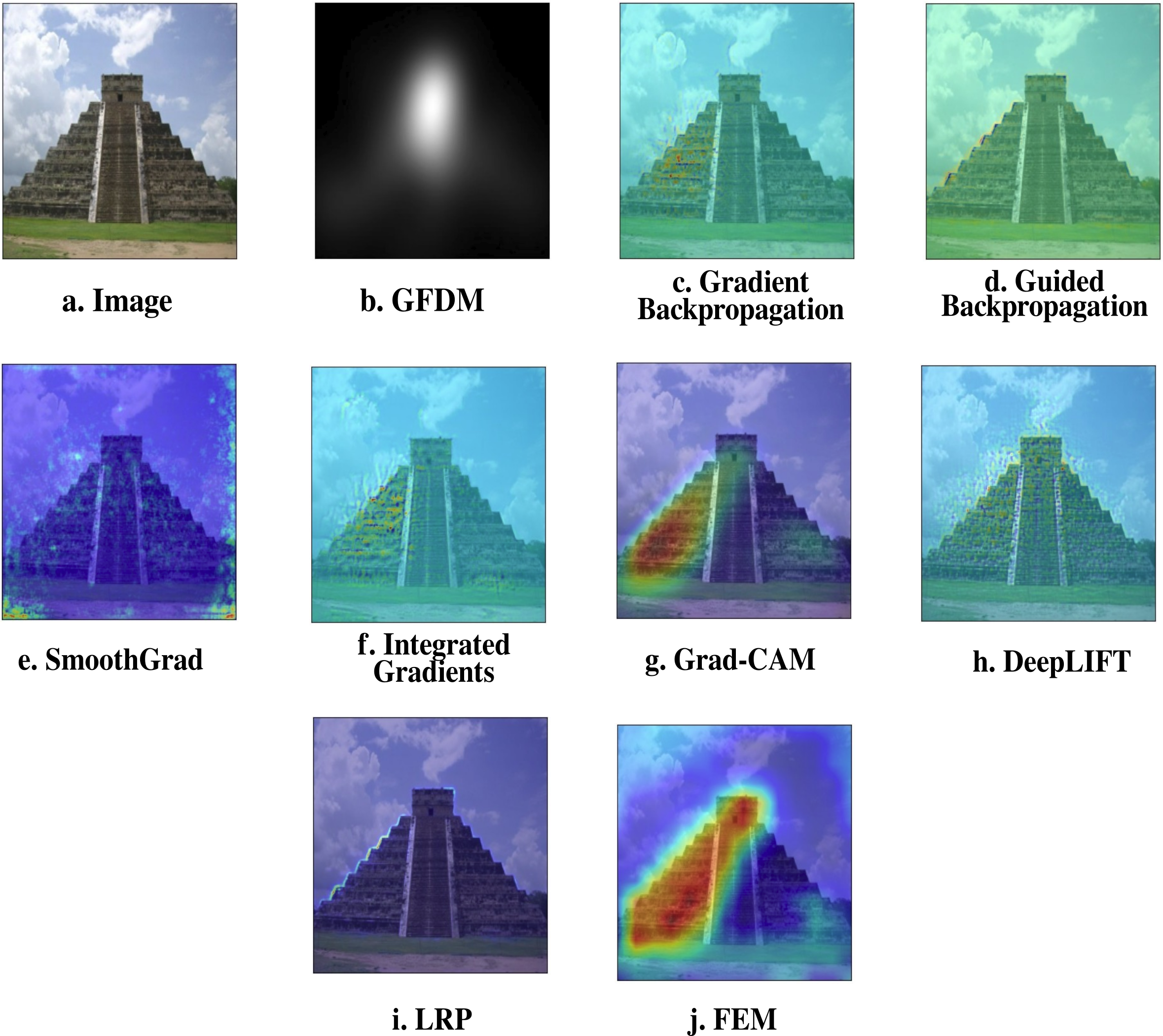}
    \caption{Heatmap visualizations for the different white-box methods for a sample image from the MexCulture dataset along with the Gaze Fixation Density Map (GFD)}
    \label{expln_maps_all}
\end{figure}

From the table, one can see that FEM is the most similar to the GFDM in SIM - L1 norm and that Grad-CAM has the highest correlation with the GFDM. The methods belonging to the linearization category (Sec.~\ref{sec:linear}) Integrated gradients and SmoothGrad are designed to reduce the noise from Gradient backpropagation heatmaps. From the table, it can be seen that the Similarity and PCC metrics are improved for these methods when compared to Gradient Backpropagation. Grad-CAM maps have the highest similarity to the GFDM maps from the linearization category of methods. Fig.~\ref{expln_maps_all} shows the heatmaps of different white-box methods for a sample image from the dataset. A zero image (black) has been taken as the reference image for the DeepLIFT visualization in the figure. It can be seen that the LRP method highlights the contrasted edges between the structure and sky to be the most relevant regions.

Muddamsetty et al.\cite{mudamsetty2020} also create a dataset comprising the user saliency maps in the form of GFDMs of medical experts on retinal images. They compare Grad-CAM and their own explainer maps with GFDM.  Metrics like Area Under the Curve (AUC) and Kullback-Leibler Divergence (KL-Div)\cite{bylinskii2018different} for the two maps were computed for the comparison. They show that the saliency map of both explainers closely aligns with human experts.

When the gaze fixation density maps are not available - and this is usually the case in real-world applications, the importance map produced by the explainer can be compared with already existing methods, which have been assessed. Thus for FEM\cite{fuad2020features} the authors compare the saliency maps obtained by FEM with gradient-based methods. They show that the most similar explanations in terms of the metrics of comparison of saliency maps such as Pearson Correlation Coefficient,  $L_1$-Similarity are given by Grad-CAM method (Sec.~ \ref{sec:Grad-CAM}) which is also observed from our results (Tab.~\ref{tab:res}). 

Methods can also be compared based on their computation time. The choice of methods is limited in their application for a larger set of images or real-time feedback if they have a long computation time. Fuad et al.\cite{fuad2020features} observe that gradient-based methods including Grad-CAM have longer computation times and FEM was faster in comparison.

\section{Conclusion}
In this review, we have attempted to provide a comprehensive overview of the current explanation methods for CNNs in image classification tasks.  After a short excursion into black-box methods which explain the decision by masking the input, we focused on white box methods as they can leverage the extra information that is available from the knowledge of the architecture of the network. 

The methods that we have discussed are focused on explaining the decision for a single input image by creating saliency maps that attribute importance scores to each of the pixels based on its contribution to the final output of the CNN. Multiple approaches have been used to calculate this contribution and based on recent works, we proposed a categorization of methods to understand similar approaches and compare their performance.

The first category we presented is the methods based on Linearization of CNNs (Sec.~\ref{sec:linear}) that approximate a CNN as a linear function and explain the decision based on gradient calculation. The deconvolution based approach and gradient backpropagation were shown to be similar except in the rectification step (ReLU operation), while the Guided backpropagation method combines the rectification steps of these methods to improve visualization. SmoothGrad and Integrated Gradients computed explanation maps of gradient backpropagation over multiple variations of the input image and combined them to produce visualizations with reduced noise. On the other hand, the Grad-CAM method computed the gradient only till the last conv layer and used an interpolation method to create the explanation method.

The second category considered were the methods based on network structure (Sec.~\ref{sec:net_arch}). LRP and DeepLIFT methods define local redistribution rules to redistribute the relevance based on the output score from the last layer to the input. DeepLIFT varies in the regard that uses a reference image and assigns relevance based on the difference in the responses of the neurons for the reference image and the input image. FEM combines the idea of Grad-CAM and uses statistical filtering to identify the strong features from the last conv layer and uses interpolation to create the final explanation map.

The final and third categories are methods based on the adversarial approach (Sec.~\ref{sec:adversary}). These methods are distinct from the other categories as they involve the use of generative networks and adversarial attack based methods to explain the important pixels in the image. Unlike the other categories, there are only a few recent methods that have employed this approach but are interesting to be explored as they give insights about the security aspect of the networks and might be used in real-world applications.

Finally, we have analysed different ways of evaluation of explanation methods: by perturbation/randomization and by comparing with human perception of images. We share the point that an explainer is good when the explanations match human intuitions contrarily to the perturbation /randomization methods which assess the explainer only from the classifier's perspective. Thus, we have conducted experience on comparison of a bunch of explainers w.r.t human gaze fixation density maps obtained from a task-driven psychovisual experiment. FEM and Grad-CAM methods were seen to be holistic and more human interpretable methods based on our experiments. Collecting these kinds of maps is often time-consuming and further methods for the evaluation of explanation methods have to be developed.

This review has been proposed from the image understanding perspective. However the explanation of AI classifiers is necessary for various kinds of imaging and signal data. The crucial aspect is the bridge between explainability and trustworthiness of AI, and this is a research question open for further contribution from the image research community.

\section{Disclosures}
The authors declare that they have no affiliations with or involvement in any
organization or entity with any financial interest  or non-financial interest in the subject matter or materials discussed in this manuscript.

\section{Acknowledgement}
This research was supported by University of Bordeaux/LaBRI.

\appendix
\section{Filtering operation in DeconvNet and Gradient calculation} \label{deconv_grad}

\begin{figure}[H]
    \centering
    \includegraphics{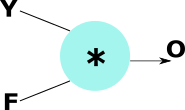}
    \caption{Convolution operation neuron in a CNN}
    \label{conv_neur}
\end{figure}

Consider a convolution neuron as shown in Fig.~\ref{conv_neur} where the operation performed is $Y*F = O$, with Y as the input, F the convolution layer filter and O the output. In a standard CNN during backpropagation of the loss $L$, the neuron receives a partial gradient of $\frac{\partial L}{\partial O}$ and the gradients to be calculated are $\frac{\partial L}{\partial Y}$ and $\frac{\partial L}{\partial F}$. 

According to the chain rule, the calculation of the partial gradient $\frac{\partial L}{\partial Y}$ is given by Eq.~(\ref{eq:chain_rule_normal}).

\begin{equation}
    \frac{\partial L}{\partial Y} = \frac{\partial L}{\partial O} \cdot \frac{\partial O}{\partial Y}
    \label{eq:chain_rule_normal}
\end{equation}
To go through a step-by-step calculation of these gradients we suppose that the input $Y$ is a matrix of $3 \times 3$, the filter F is a matrix of size $2 \times 2$ and the convolution operation is the one with a stride of $1$. Then the corresponding matrices and the loss gradient that is backpropagated from the following layer would be as shown in Eq.~(\ref{eq:matrices}).

 \begin{equation}
Y =  \begin{bmatrix}
y_{11} & y_{12}  & y_{13} \\ 
y_{21} & y_{22} & y_{23} \\ 
 y_{31}& y_{32} & y_{33}
\end{bmatrix}
,
F = \begin{bmatrix}
f_{11} & f_{12}  \\ 
f_{21} & f_{22}  \\ 
\end{bmatrix}
 ,
 O = \begin{bmatrix}
o_{11} & o_{12}  \\ 
o_{21} & o_{22}  \\ 
\end{bmatrix}
, 
 \frac{\partial L }{\partial O} = \begin{bmatrix}
\frac{\partial L}{ \partial o_{11} } & \frac{\partial L}{ \partial o_{12} } \\ 
\frac{\partial L}{ \partial o_{21} } & \frac{\partial L}{ \partial o_{22} }   \\ 
\end{bmatrix} 
\label{eq:matrices}
\end{equation}

The equations for the calculation of the convolution during the forward pass yields the expressions of the $O$ as shown in Eq. ~(\ref{eq:calc_output}).
\begin{equation}
    \begin{split}
        o_{11} = y_{11} f_{11} +  y_{12} f_{12} +  y_{21} f_{21} +  y_{22} f_{22} \\
    o_{12} = y_{12} f_{11} +  y_{13} f_{12} +  y_{22} f_{21} +  y_{23} f_{22} \\
    o_{21} = y_{21} f_{11} +  y_{22} f_{12} +  y_{31} f_{21} +  y_{32} f_{22} \\
    o_{22} = y_{22} f_{11} +  y_{23} f_{12} +  y_{32} f_{21} +  y_{33} f_{22}
    \label{eq:calc_output}
    \end{split}
\end{equation}

So to calculate the partial gradient of the loss w.r.t. the input, the first calculation that needs to be done is $\frac{\partial O}{\partial Y}$. A single calculation of this expression is shown in Eq. ~(\ref{eq:partial_o}) based on the Eqs.~(\ref{eq:calc_output}). The rest of the terms can be calculated similarly.
 
 \begin{equation}
 \frac{\partial o_{11}}{\partial y_{11}} = f_{11}
 ,
 \frac{\partial o_{11}}{\partial y_{12}} = f_{12}
 ,
 \frac{\partial o_{11}}{\partial y_{21}} = f_{21}
 ,
 \frac{\partial o_{11}}{\partial y_{22}} = f_{22}
 \label{eq:partial_o}
 \end{equation}

Subsequently, the partial gradient of the loss w.r.t the input would be given by Eqs.~(\ref{eq:loss_input_grad}).

\begin{equation}
    \begin{split}
        &\frac{\partial L}{\partial y_{11}} = \frac{\partial L}{\partial o_{11}} \cdot f_{11}, \hspace{0.5cm} \frac{\partial L}{\partial y_{12}} = \frac{\partial L}{\partial o_{11}} \cdot f_{12} + \frac{\partial L}{\partial o_{12}} \cdot f_{11}, \hspace{0.5cm}  \frac{\partial L}{\partial y_{13}} = \frac{\partial L}{\partial o_{12}} \cdot f_{12} \\
        &\frac{\partial L}{\partial y_{21}} = \frac{\partial L}{\partial o_{11}} \cdot f_{21} + \frac{\partial L}{\partial o_{21}} \cdot f_{11},  \hspace{0.5cm}  \frac{\partial L}{\partial y_{22}} = \frac{\partial L}{\partial o_{11}} \cdot f_{22} + \frac{\partial L}{\partial o_{12}} \cdot f_{21} + \frac{\partial L}{\partial o_{21}} \cdot f_{12} + \frac{\partial L}{\partial o_{22}} \cdot f_{11} \\
       &\frac{\partial L}{\partial y_{23}} = \frac{\partial L}{\partial o_{12}} \cdot f_{22} + \frac{\partial L}{\partial o_{22}} \cdot f_{12},  \hspace{0.5cm}  \frac{\partial L}{\partial y_{31}} = \frac{\partial L}{\partial o_{21}} \cdot f_{21}\\
       &\frac{\partial L}{\partial y_{32}} = \frac{\partial L}{\partial o_{21}} \cdot f_{22} + \frac{\partial L}{\partial o_{22}} \cdot f_{21},  \hspace{0.5cm}  \frac{\partial L}{\partial y_{33}} = \frac{\partial L}{\partial o_{22}} \cdot f_{22}\\
    \end{split}
    \label{eq:loss_input_grad}
\end{equation}

Thus the partial gradient $\frac{\partial L}{\partial Y}$ when calculated using the chain rule can be written as a full convolution (when the loss matrix is zero-padded to have full convolution operation) of the $180^{\circ}$ inverted filter, i.e. the filter matrix has been flipped vertically and then horizontally as shown in Eq.~(\ref{eq:filter_backprop}), and the loss gradient matrix $\frac{\partial L}{\partial O}$.

\begin{equation}
    F = \begin{bmatrix}
f_{22} & f_{21}  \\ 
f_{12} & f_{11}  \\ 
\end{bmatrix}
\label{eq:filter_backprop}
\end{equation}

DeconvNet (Sec.~\ref{sec:deconv_net}) uses the same operation at the filtering step. Thus, the deconvolution step of the DeconvNet and the calculation of the gradient with respect to the input at a convolution layer are equivalent.

\bibliography{report}   
\bibliographystyle{spiejour}   


\vspace{2ex}\noindent\textbf{Meghna P Ayyar} has completed her MS in Image Processing and Computer Vision (IPCV) and is currently working as an Associate Junior Researcher at LaBRI, Bordeaux. Her topics of interest include computer vision, Explainable AI and applications of deep learning to healthcare.

\vspace{2ex}\noindent\textbf{Jenny Benois-Pineau} is a full professor of Computer Science at the University Bordeaux. Her topics of interest include image/multimedia, artificial intelligence. She has authored and co-authored more than 220 papers and is an associated editor of SPIC, ACM MTAP, senior associate editor JEI SPIE journals. She has Knight of Academic Palms grade and has served in program committees of major international conferences in her field. She is a member of IEEE IVMSP TC committee.  

\vspace{2ex}\noindent\textbf{Akka Zemmari} is a full professor of Computer Science at the University of Bordeaux. His research interests include AI, Deep learning, distributed algorithms and systems, graphs, randomized algorithms and security. He is co-coordinator of AI research axis in LaBRI and has been a researcher in several National, European and Europe-India research projects. He is the author and co-author of more than 80 research papers, conference proceedings and book chapters.


\listoffigures
\listoftables

\end{spacing}
\end{document}